# Discovering Support and Affiliated Features from Very High Dimensions

**Yiteng Zhai** YZHAI1@NTU.EDU.SG
**Mingkui Tan** TANM0097@NTU.EDU.SG
**Ivor W. Tsang** IVORTSANG@NTU.EDU.SG
**Yew-Soon Ong** ASYSONG@NTU.EDU.SG
Centre for Computational Intelligence ($C^2i$), Nanyang Technological University (NTU), Singapore

## Abstract

In this paper, a novel learning paradigm is presented to automatically identify groups of informative and correlated features from very high dimensions. Specifically, we explicitly incorporate correlation measures as constraints and then propose an efficient embedded feature selection method using recently developed cutting plane strategy. The benefits of the proposed algorithm are two-folds. First, it can identify the optimal discriminative and uncorrelated feature subset to the output labels, denoted here as **Support Features**, which brings about significant improvements in prediction performance over other state of the art feature selection methods considered in the paper. Second, during the learning process, the underlying group structures of correlated features associated with each support feature, denoted as **Affiliated Features**, can also be discovered without any additional cost. These affiliated features serve to improve the interpretations on the learning tasks. Extensive empirical studies on both synthetic and very high dimensional real-world datasets verify the validity and efficiency of the proposed method.

## 1. Introduction

Many real-world datasets in text and digital media domains are typically represented with very high dimensional features, bringing significant challenges in data mining. Learning performance is often degraded with inflating of dimensions, leading to the well-known notion of “curse of dimensionality”. This problem becomes particularly critical when the number of informative features is relatively small, but involved with a vast variety of irrelevant features and redundant features (Yu & Liu, 2004).

To address this issue, a plethora of feature selection methods have been developed in the recent decades. In general, these methods have been categorized as three core themes (Guyon, 2008): filter methods (Yu & Liu, 2003; Peng et al., 2005), wrapper methods (Guyon & Elisseeff, 2003; Zhu et al., 2007) and embedded methods (Yuan et al., 2011; Tan et al., 2010; Mao & Tsang, 2011). Specifically, filter methods select informative features based on their individual discriminative power or correlation criterion. The benefits of filter methods lie in their low computational requirements. The drawback however is that it may not identify the optimal feature subset suitable for the predictive model of interest. On the contrary, since wrapper methods, such as SVM-RFE (Guyon & Elisseeff, 2003), select the discriminative features solely based on the inductive learning rule, they typically exhibit higher predictive performance but at the expense of a lower computational efficiency on large scale and very high dimensional problems. Embedded methods refer to approaches that directly optimize some regularized risk function w.r.t. two sets of parameters: parameter of the learning machine, and parameter to control the feature sparsity (Guyon, 2008). As such they are usually more efficient than wrapper methods.

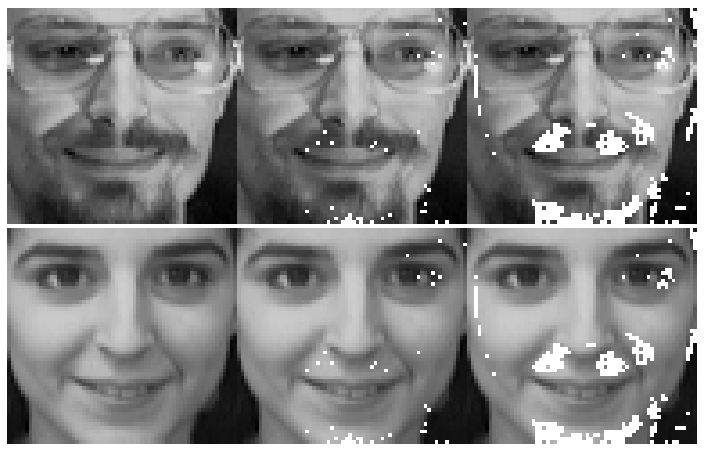

*Figure 1.* ORL face dataset (Left: original face images; Middle: face images with selected support features (white pixels); Right: face images with selected affiliated features (white pixels).

To date, it is worth noting that most methods in all three categories generally assume a good feature subset has strong prediction ability pertaining to the output labels; meanwhile the selected features should also maintain low correlations. In other words, correlated features are deemed as redundant, and this redundancy should be minimized (Hall, 1999; Guyon, 2008; Zhao et al., 2011; 2012). Though elim-inating redundant features has been widely used in practice and regarded as the guiding principle behind the development of feature selection methods, it may not always hold since these correlated features can be useful and informative for the tasks on hand. As also discussed in (O'Sullivan et al., 2000; Caruana & de Sa, 2003; Xu et al., 2012), such feature redundancy has the benefits of bringing about stable generalization performances. Here, we present an illustrating example using Figure 1. Despite a perfect prediction by the classifier on the male and female face images based on the optimal features identified (denoted here as **support features**), these uncorrelated features which are depicted by the white pixels in the middle of Figure 1, can be observed to be sparsely spread across the entire image, thus giving little cue to assist the human user in the interpretation of the features. The white pixels shown in the right column of Figure 1, denoting the highly correlated features, on the other hand, are much more informative. To be precise, it is easy for the human user to spot the core group features in the regions of the mustache and beard, which can be helpful to the users in grasping a better understanding on the critical objects in the images so that further analysis can be made. On the contrary, existing feature selection methods usually eliminate these correlated features and treat them as pure redundancies. Further, the act of identifying these correlated features from high dimensions is typically very compute intensive.

Taking this cue, in this paper, we introduce an efficient feature selection method that can identify the optimal feature subset to the output labels; while minimizing the correlation among the selected features. Each selected feature is defined as **Support Feature**; while the correlated features associate with this support feature are denoted as **Affiliated Features**, and are discovered during the feature selection process without any additional cost. The core contributions of this paper are listed as follows:

1. We develop an efficient Correlation Redundancy Matching (CRM) algorithm, which accommodates the correlation constraints among features, to identify both discriminative features and correlated features. Similar to L1-SVM, the proposed method can be categorized as an embedded feature selection approach and attains global convergence. Theoretical analysis and empirical studies show that our method is scalable to very high dimensional problems.
2. For each support feature, the respective associated affiliated feature subset contains correlated features that are informative to the output labels and further, with these two types of features, a group structure of features can be established. These affiliated groups maintain redundancy for robust prediction, and help users understand the learning tasks for further analysis.
3. Empirical studies on both synthetic and real-world experiments verify the superior performance of the proposed method over the state-of-the-art feature selection methods in terms of both testing accuracies and redundancy rate (Zhao et al., 2012).

## 2. Preliminaries and Related Works

In this paper, we denote a data point by $\mathbf{x}_i \in \mathbb{R}^n$ and the dataset by $\mathbf{X} = [\mathbf{x}_1, \dots, \mathbf{x}_n] = [\mathbf{f}_1', \dots, \mathbf{f}_m']' \in \mathbb{R}^{m \times n}$, where $\mathbf{f}_j$ represents a row vector corresponding to the $j^{th}$ feature of the data points in $\mathbf{X}$. Each $\mathbf{x}_i$ is associated with an output $y_i \in \{\pm 1\}$. We also define $\mathbf{y}$ as the vector of the label for the data. Symbols $\mathbf{0}$ and $\mathbf{1}$ are the column vectors with all zeros and all ones, respectively. For any vector $\mathbf{f}$, we denote the mean and standard deviation of the entries in $\mathbf{f}$ as $\mu_\mathbf{f}$ and $\sigma_\mathbf{f}$, respectively. Additionally, we denote the element wise product between two matrices $\mathbf{A}$ and $\mathbf{B}$ by $\mathbf{A} \odot \mathbf{B}$. Finally, we denote $|\mathcal{S}|$ to be the size of a set $\mathcal{S}$.

As previously stated, feature correlation is of particular interest in our present research. In the past decades, a large branch of feature selection approaches have targeted on reducing the redundancy among the selected features. The notion of feature redundancy is usually measured by means of feature correlation. The major motivation has been to find the optimal or minimized feature subset corresponding to the output labels. Thus, when a feature is selected, other features that are highly related to this feature is typically rejected so as to minimize feature redundancies.

### 2.1. Feature Correlation Measures

To date, various criteria have been introduced for defining the correlation between features. For instance, a widely used correlation criterion is the Pearson's correlation coefficient (PCC), which measures the linear correlation between variables. In more details, given two feature vectors $\mathbf{f}_j$ and $\mathbf{f}_k$, the metric $\rho$ in PCC can be defined as follows,

$$\rho(\mathbf{f}_j, \mathbf{f}_k) = \frac{\mathbf{cov}(\mathbf{f}_j, \mathbf{f}_k)}{\sigma_{\mathbf{f}_j}\sigma_{\mathbf{f}_k}} = \frac{\frac{1}{n}(\mathbf{f}_j - \mu_{\mathbf{f}_j}\mathbf{1}')(\mathbf{f}_k - \mu_{\mathbf{f}_k}\mathbf{1}')'}{\sigma_{\mathbf{f}_j}\sigma_{\mathbf{f}_k}}. \quad (1)$$

Notice that $\rho$ is a symmetrical measure that ranges in $[-1, 1]$. If two variables are fully independent, $\rho = 0$. On the other hand, when the two variables are completely correlated to each other, namely, one variable can exactly predict another variable, we have $|\rho| = 1$. Other metrics such as information gain $\mathrm{IG}(\mathbf{f}_j|\mathbf{f}_k)$ and symmetrical uncertainty $\mathrm{SU}(\mathbf{f}_j|\mathbf{f}_k)$ have also been discussed in (Yu & Liu, 2003).

### 2.2. Feature Redundancy Reduction

Based on these correlation measures, several methods have attempted to reduce the redundancy among the selected features. For instance, in Fast Correlation Based Filter (FCBF) (Yu & Liu, 2003), feature importance and feature correlation are assessed by means of SU measure. FCBF first selects a set of predominant features that are relevant to the output labels. Subsequently, the informative features to labels are kept while the correlated features are removed based on some elegantly designed intuitive rules (Yu & Liu, 2003). Another notable redundancy reduction method is Minimum Redundancy Maximum Relevance (mRMR) (Peng et al., 2005), which selects the most correlated features to the labels such that they are mutually far apart from each other by maximizing the dependency between the joint distribution of the selected features and the output labels.

Zhou *et al.* (2010) proposed Redundancy Constrained Feature Selection (RCFS), which first performs feature clustering by using some distance measures (e.g., $1 - |\rho(\mathbf{f}_j, \mathbf{f}_k)|$). Hence the correlated features may be grouped into several clusters. After that, some features are then identified from each cluster. Next, a feature subset is further identified from the selected features in each cluster group using graph based feature selection criteria (Nie et al., 2008) that capture the global and local intrinsic structures of the data. This strategy however is heavily sensitive to the choice of graph Laplacian matrices used. For example, the Laplacian score is usually constructed using $K$ nearest neighbor ($K$NN). In practice, on very high dimensional problems, $K$NNs can be very far away from each other in reality due to the effect of the curse of dimensionality. Besides, the high computational cost of feature clustering on high dimensional data and graph based methods (taking $O(n^2 m)$) make this approach less attractive on large scale data. Recently, Zhao *et al.* (2012) proposed a framework to unify different criteria for removing feature redundancies. Nevertheless, existing methods have remained to focus on reducing these redundancies. More importantly, to date the discovery of correlated yet informative features has been relatively unexplored.

## 3. Group Discovery Machine

In this section, we present an efficient automatic feature grouping method, which identifies groups of discriminative yet correlated features. Similar to (Guyon, 2008), a vector $\boldsymbol{\delta} = [\delta_1, \ldots, \delta_m]' \in \{0, 1\}^m$ is introduced to indicate whether the corresponding feature is selected ($\delta_j = 1$) or not ($\delta_j = 0$), such that the decision function is defined as: $f(\mathbf{x}) = \mathbf{w}'(\mathbf{x} \odot \boldsymbol{\delta})$, where $\mathbf{w} = [w_1, \cdots, w_m]'$ is the weight vector. To limit the number of selected features to be less than $B$, the $\ell_0$ constraint $\|\boldsymbol{\delta}\|_0 \leq B$ is imposed for the purpose of feature selection (Nie et al., 2008).

### 3.1. Correlation Constraints

To control the correlation among the selected features, we explicitly introduce the following constraint on $\boldsymbol{\delta}$ such that,

$$\delta_j \delta_k = 0, |\rho(\mathbf{f}_j, \mathbf{f}_k)| \geq 1 - \tau, \forall j \neq k, \tag{2}$$

which states that any selected feature pair should not be correlated so long as their coefficient defined in (1) does not exceeds $(1 - \tau)$ where $\tau \in (0, 1)$. We also define $\boldsymbol{\Delta} = \{\boldsymbol{\delta} | \sum_{j=1}^m \delta_j \leq B;\ \delta_j \in \{0, 1\},\ \forall j = 1, \cdots, m;\ \delta_j \delta_k = 0, |\rho(\mathbf{f}_j, \mathbf{f}_k)| \geq 1 - \tau, \forall j \neq k\}$ as the domain of $\boldsymbol{\delta}$. Here, (2) defines $O(m^2)$ quadratic constraints with $m$ integer variables, thus finding the solution $\boldsymbol{\delta} \in \boldsymbol{\Delta}$ involves combinatorial subset selection, resulting in high computational cost especially when the dimension $m$ is high.

### 3.2. Proposed Formulation

Here, we aim to find a large margin decision function $f(\mathbf{x})$ for robust prediction, and seamlessly identify the informative yet uncorrelated feature subset that satisfies the constraint in (2). For the sake of simplicity, we use the square hinge loss in SVM, and arrive at the following problem:

$$\begin{aligned} \min_{\boldsymbol{\delta} \in \boldsymbol{\Delta}} \min_{\mathbf{w}, \gamma, \xi} \quad & \frac{1}{2}\|\mathbf{w}\|_2^2 - \gamma + \frac{C}{2}\sum_{i=1}^n \xi_i^2 \\ s.t. \quad & y_i \mathbf{w}'(\mathbf{x}_i \odot \boldsymbol{\delta}) \geq \gamma - \xi_i, \ \ i = 1, \cdots, n, \end{aligned} \tag{3}$$

where $\xi_i \geq 0$ is the slack variable, $\gamma / \|\mathbf{w}\|$ denotes the margin and $C$ is a tradeoff parameter to regulate the function complexity $\|\mathbf{w}\|_2^2$ and the training error ($\xi_i$'s). Note, as discussed in Section 3.1, the optimization problem in (3) with the constraints defined in $\boldsymbol{\Delta}$ is very challenging.

### 3.3. Cutting Plane Algorithm

To tackle this, we first transform the inner minimization in (3) w.r.t. $\mathbf{w}, \gamma, \xi_i$ into the dual of SVM, then (3) becomes a minimax saddle-point problem. Inspired by (Tan et al., 2010), by applying the minimax optimization theory, one can obtain a tight convex relaxation to (3), which is in the form of the following Quadratically Constrained Quadratic Programming (QCQP) problem:

$$\min_{\boldsymbol{\alpha} \in \mathcal{A}, \theta} \theta : \theta \geq g_{\boldsymbol{\delta}}(\boldsymbol{\alpha}),\ \forall \boldsymbol{\delta} \in \boldsymbol{\Delta} \quad \text{or} \quad \min_{\boldsymbol{\alpha} \in \mathcal{A}} \max_{\boldsymbol{\delta} \in \boldsymbol{\Delta}} g_{\boldsymbol{\delta}}(\boldsymbol{\alpha}), \tag{4}$$

where $g_{\boldsymbol{\delta}}(\boldsymbol{\alpha}) = \frac{1}{2}\left\| \sum_{i=1}^n \alpha_i y_i (\mathbf{x}_i \odot \boldsymbol{\delta}) \right\|^2 + \frac{1}{2C}\boldsymbol{\alpha}'\boldsymbol{\alpha}$, $\boldsymbol{\alpha} = [\alpha_1, \ldots, \alpha_n]'$ is the vector of dual variables, $\mathcal{A} = \{\boldsymbol{\alpha} | \sum_{i=1}^n \alpha_i = 1, \alpha_i \geq 0, \forall i = 1, \cdots, n\}$ is the domain of $\boldsymbol{\alpha}$, and $\theta$ is the upper bound of $g_{\boldsymbol{\delta}}(\cdot)$. Nevertheless, since there are as many as $(\sum_{i=0}^B \binom{m}{i})$ quadratic constraints in (4), it remains computationally expensive to solve (4). Rather than solving the original problem with a large collection of constraints, the cutting plane strategy (Mutapcic & Boyd, 2009) can be employed to iteratively generate a set of active constraints and then solve this reduced optimization problem with the current constraint set. Since $\max_{\boldsymbol{\delta} \in \boldsymbol{\Delta}} g_{\boldsymbol{\delta}}(\boldsymbol{\alpha}) \geq g_{\boldsymbol{\delta}^t}(\boldsymbol{\alpha}), \forall \boldsymbol{\delta}^t \in \boldsymbol{\Delta}$, with a reduced active constraint set $\mathcal{C} \subset \boldsymbol{\Delta}$, the lower bound approximation of (4) can be obtained by $\max_{\boldsymbol{\delta} \in \boldsymbol{\Delta}} g_{\boldsymbol{\delta}}(\boldsymbol{\alpha}) \geq \max_{t=1,\ldots,T} g_{\boldsymbol{\delta}^t}(\boldsymbol{\alpha})$ with $T = |\mathcal{C}|$, where $T$ is the maximum number of constraints that will be added. This leads to solving a reduced problem of (4) as follows,

$$\min_{\boldsymbol{\alpha} \in \mathcal{A}, \theta} \theta \ : \ \theta \geq g_{\boldsymbol{\delta}^t}(\boldsymbol{\alpha}), \ \forall\, \boldsymbol{\delta}^t \in \mathcal{C}. \tag{5}$$

The details to solve (5) are outlined in Algorithm 1, where some notations will be explained later. Specifically, at each

**Algorithm 1** Group Discovery Machine

**Input:** Given a dataset $(\mathbf{X}, \mathbf{y})$, parameter $B$ and $\tau$.
**Output:** $\mathcal{S}$ and $\mathcal{Q}$ are the index sets of support features and affiliated features, respectively.
Set $\boldsymbol{\alpha} = \mathbf{1}/n, \mathcal{S} = \emptyset$ and $\mathcal{Q} = \emptyset$.
**for** $t = 1$ **to** $T$ **do**
  1: Call $\boldsymbol{\delta}^t$ = CRM$(\mathbf{X}, \mathbf{y}, B, \tau, \boldsymbol{\alpha}, \mathcal{S}, \mathcal{Q})$ to find the most violated $\boldsymbol{\delta}^t$ and acquire $\mathcal{S}, \mathcal{Q}$
  2: Set $\mathcal{C} = \mathcal{C} \cup \{\boldsymbol{\delta}^t\}$
  3: Solve (5) defined on $\mathcal{C}$ while updating $\boldsymbol{\alpha}$
**end for**

**Algorithm 2** CRM$(\mathbf{X}, \mathbf{y}, B, \tau, \boldsymbol{\alpha}, \mathcal{S}, \mathcal{Q})$

**Input:** Given a dataset $(\mathbf{X}, \mathbf{y})$, parameter $B$ and $\tau$, $\mathcal{S}$ and $\mathcal{Q}$ are the index sets of support features and affiliated features, respectively. Initialize an index set $\mathcal{B} = \emptyset$.
**Output:** A zero-one vector $\boldsymbol{\delta} \in \mathbb{R}^m$, initialized as $\boldsymbol{\delta} = \mathbf{0}^m$.
1: Compute $\mathbf{c} = \sum_{i=1}^n \alpha_i y_i \mathbf{x}_i$, sort $|c_j|$ in the descending order, and record the feature ranking list as $\mathcal{E}$.
2: $\mathcal{G} = \emptyset$ denotes a temporary affiliated feature set and $k = 1$.
**while** $|\mathcal{B}| \le B$ **do**
  Pick the $k^{th}$ feature $\mathbf{f}_z$ from $\mathbf{X}$, where $z = \mathcal{E}(k)$
  Set $\mathcal{B} = \mathcal{B} \cup \{z\}$ and $\mathcal{G} = \mathcal{G} \cup \{z\}$
  **while** $k < m$ **do**
    $k = k + 1$
    Pick the current $k^{th}$ feature $\mathbf{f}_h$ from $\mathbf{X}$, where $h = \mathcal{E}(k)$
    **if** $(|\rho(\mathbf{f}_z, \mathbf{f}_h)| \ge 1 - \tau)$ **then**
      Update $\mathcal{G} = \mathcal{G} \cup \{h\}$
    **end if**
    **if** $(|c_z| - \sqrt{2\tau}\|\boldsymbol{\alpha}\| > |c_h|)$ **then**
      **break**
    **end if**
  **end while**
  Set $\mathcal{Q} = \mathcal{Q} \cup \{\mathcal{G}\}$ and $\mathcal{G} = \emptyset$.
**end while**
Update $\mathcal{S} = \mathcal{S} \cup \{\mathcal{B}\}$ and $\boldsymbol{\delta}_{\mathcal{B}} = \mathbf{1}$.

iteration of Algorithm 1, one needs to solve the worst case analysis (the same as finding the most violated constraint $\boldsymbol{\delta}^t$) of Problem (4), which shall be described in Section 3.4. Subsequently, the obtained $\boldsymbol{\delta}^t$ would be appended into the active constraint set $\mathcal{C}$. Finally, the problem w.r.t. a reduced active constraint set $\mathcal{C}$ can be solved by some efficient QCQP solvers (Tan et al., 2010).

To summarize, the cutting plane algorithm generally converges to a robust optimal solution within tens of iterations with the exact worst case analysis and shows good performance in many real applications (Mutapcic & Boyd, 2009).

### 3.4. Correlation Redundancy Matching

In this subsection, we discuss the worst case analysis of problem (4) (i.e., equivalent to finding the most violated constraints), which plays the key role in cutting plane algorithms (Mutapcic & Boyd, 2009). In our setting, this translate to solving the following integer optimization problem:

$$\max_{\boldsymbol{\delta} \in \boldsymbol{\Delta}} \quad \Big\| \sum_{i=1}^n \alpha_i y_i (\mathbf{x}_i \odot \boldsymbol{\delta}) \Big\|^2. \qquad (6)$$

In general, solving this problem is NP hard. However, since $\frac{1}{2}\| \sum_{i=1}^n \alpha_i y_i (\mathbf{x}_i \odot \boldsymbol{\delta})\|^2 = \frac{1}{2}\| \sum_{i=1}^n (\alpha_i y_i \mathbf{x}_i) \odot \boldsymbol{\delta}\|^2 = \frac{1}{2}\sum_{j=1}^m c_j^2 \delta_j$, where $c_j = \sum_{i=1}^n \alpha_i y_i x_{ij} = \mathbf{f}_j \tilde{\boldsymbol{\alpha}}$, and $\tilde{\boldsymbol{\alpha}} = [\alpha_1 y_1, \ldots, \alpha_n y_n]'$, this indicates that the informative features accord with the features with the highest $|c_j|$ 's. In addition, based on this observation, the following proposition[1] will further show that for a set of correlated features, if one of them is informative to the output labels, all of them can be deem as informative to the output labels as well.

**Proposition 1.** *Given a nonzero column vector $\tilde{\boldsymbol{\alpha}}$, and any two feature vectors $\mathbf{f}_1$ and $\mathbf{f}_2$ that $\sigma_{\mathbf{f}_1} = \sigma_{\mathbf{f}_2} = 1/\sqrt{n}$ and $\mu_{\mathbf{f}_1} = \mu_{\mathbf{f}_2} = 0$. Suppose $|\rho(\mathbf{f}_1, \mathbf{f}_2)| \ge 1 - \tau$, then $|\,|\mathbf{f}_1\tilde{\boldsymbol{\alpha}}| - |\mathbf{f}_2\tilde{\boldsymbol{\alpha}}|\,| \le \sqrt{2\tau}\|\tilde{\boldsymbol{\alpha}}\|$, where $\tau \in (0, 1)$.*

*Proof.* With $|\rho(\mathbf{f}_1, \mathbf{f}_2)| \ge 1 - \tau$, using (1), we have $|\rho(\mathbf{f}_1, \mathbf{f}_2)| = |\mathbf{f}_1\mathbf{f}_2'| \ge 1 - \tau$, namely, $\mathbf{f}_1\mathbf{f}_2' \ge 1 - \tau$ (positive correlation) or $\mathbf{f}_1\mathbf{f}_2' \le \tau - 1$ (negative correlation). Suppose $\mathbf{f}_1$ and $\mathbf{f}_2$ are positive correlated, we have $\|\mathbf{f}_1 - \mathbf{f}_2\|^2 = \|\mathbf{f}_1\|^2 + \|\mathbf{f}_2\|^2 - 2\mathbf{f}_1\mathbf{f}_2' = 2(1 - \mathbf{f}_1\mathbf{f}_2') \le 2\tau$, as $\|\mathbf{f}_1\|^2 = \|\mathbf{f}_2\|^2 = 1$ when $\sigma_{\mathbf{f}_1}^2 = \sigma_{\mathbf{f}_1}^2 = 1/n$. Note that $|\,\|\mathbf{f}_2 - \tilde{\boldsymbol{\alpha}}'\|^2 - \|\mathbf{f}_1 - \tilde{\boldsymbol{\alpha}}'\|^2\,| = |2(\mathbf{f}_1 - \mathbf{f}_2)\tilde{\boldsymbol{\alpha}}| \le 2\|\tilde{\boldsymbol{\alpha}}\|\|\mathbf{f}_1 - \mathbf{f}_2\| \le 2\sqrt{2\tau}\|\tilde{\boldsymbol{\alpha}}\|$. In other words, $|\mathbf{f}_1\tilde{\boldsymbol{\alpha}} - \mathbf{f}_2\tilde{\boldsymbol{\alpha}}| \le \sqrt{2\tau}\|\tilde{\boldsymbol{\alpha}}\|$. Hence, we have $|\,|\mathbf{f}_1\tilde{\boldsymbol{\alpha}}| - |\mathbf{f}_2\tilde{\boldsymbol{\alpha}}|\,| \le \sqrt{2\tau}\|\tilde{\boldsymbol{\alpha}}\|$. In the case of negative correlation, we define a positive correlated vector $\hat{\mathbf{f}}_2 = -\mathbf{f}_2$ and the proof follows the derivation of the positive correlation case, we complete the proof. □

The above results state that if two feature vectors $\mathbf{f}_1$ and $\mathbf{f}_2$ are highly correlated, their distance (or correlation) to any exemplar vector $\tilde{\boldsymbol{\alpha}}'$ will be very similar to one other. Then a natural question arises, considering the correlated features, which feature poses greater importance to the output labels for a given $\tilde{\boldsymbol{\alpha}}$? To address this question, we first offer the definitions of **Support Features** and **Affiliated Features**. In particular, support features refer to informative features with relatively low correlations. affiliated features, on the other hand, refer to the correlated features associated with each support feature.

**Definition 1. Support and Affiliated features:** *Given any exemplar vector $\tilde{\boldsymbol{\alpha}} \in \mathbb{R}^n$ and a collection of feature vectors $\{\mathbf{f}_i\}$, where $\mathbf{f}_i' \in \mathbb{R}^n$. The support feature is given by $\max_i |\mathbf{f}_i\tilde{\boldsymbol{\alpha}}|$ for the given $\tilde{\boldsymbol{\alpha}}$. The remaining correlated features in $\{\mathbf{f}_j\}$ w.r.t. $\mathbf{f}_i$ denote the affiliated features.*

For the sake of conciseness, we let $\mathcal{S}$ be the index set of the support features and introduce a data structure $\mathcal{Q} = \{\mathcal{G}_i\}$ to represent the hierarchical structure of features, where $\mathcal{G}_i$ denotes the index set of the **affiliated features** for the $i^{th}$ support feature. Notice that the support feature is correlated to itself, we have $\mathcal{S} \subset \mathcal{Q}$. By taking this scheme, we can keep all the correlated features rather than omitting them. Based on these definitions, once a support feature is identified (i.e., the feature with the largest $c_j$), all relevant features that correlate with this support feature will form the corresponding affiliated feature group or cluster. Since the proposed method can discover the correlated feature groups, we name it Group Discovery Machine (GDM). Note that, alternatively, one could use a brute-force approach to search across all features and identify all features that are correlated to the support feature as the affiliated features to achieve the same goal. However, such a strategy can be computationally infeasible. Fortunately, we show in what follows a theorem to illustrate that in practice one can address this problem by scanning only a small subset of the features on very high dimensional problems.

[1] Here, only the linear correlation is considered.

**Theorem 1.** *Given a nonzero column vector $\tilde{\boldsymbol{\alpha}}$ and any two feature vectors $\mathbf{f}_j$ and $\mathbf{f}_k$ that $\mu_{\mathbf{f}_j} = \mu_{\mathbf{f}_k} = 0$ and $\sigma_{\mathbf{f}_j} = \sigma_{\mathbf{f}_k} = 1/\sqrt{n}$, if $|\rho(\mathbf{f}_j, \mathbf{f}_k)| \geq 1 - \tau$ and $\mathbf{f}_j$ is the support feature with score $|c_j| = |\mathbf{f}_j \tilde{\boldsymbol{\alpha}}|$, then the score of the feature $\mathbf{f}_k$ will satisfy $|c_k| = |\mathbf{f}_k \tilde{\boldsymbol{\alpha}}| \geq |c_j| - \sqrt{2\tau}\|\tilde{\boldsymbol{\alpha}}\|$.*

*Proof.* From Proposition 1, we know that $|\,|\mathbf{f}_j \tilde{\boldsymbol{\alpha}}| - |\mathbf{f}_k \tilde{\boldsymbol{\alpha}}|\,| \leq \sqrt{2\tau}\|\tilde{\boldsymbol{\alpha}}\|$, if $|\rho| > 1 - \tau$. Since $\mathbf{f}_j$ is the support feature, so $|\mathbf{f}_j \tilde{\boldsymbol{\alpha}}| \geq |\mathbf{f}_k \tilde{\boldsymbol{\alpha}}|$, we have $|\mathbf{f}_k \tilde{\boldsymbol{\alpha}}| \geq |\mathbf{f}_j \tilde{\boldsymbol{\alpha}}| - \sqrt{2\tau}\|\tilde{\boldsymbol{\alpha}}\|$. This completes the proof. □

The above theorem says that if two features are highly correlated, their scores will be very close. In other words, for a given support feature $\mathbf{f}_j$, the feature with a score lower than $|c_j| - \sqrt{2\tau}\|\tilde{\boldsymbol{\alpha}}\|$ shall not be considered as an affiliated feature to $\mathbf{f}_j$, at the correlation level of $(1 - \tau)$. Based on this, the worst case analysis can be conducted in Algorithm 2, which is termed here as Correlation Redundancy Matching (CRM). The basic idea of CRM is that, for every iteration, we first find the support features with larger feature scores. Once the support features are identified, the corresponding affiliated features are then identified from the remaining unselected features. The whole procedure is repeated until a maximum of $B$ numbers of support features (see (2)) is selected. Thus, the computational cost of this worst case analysis can be substantially reduced, which we will go through details in the upcoming subsection.

**Proposition 2.** *With the* ***Correlation Redundancy Matching*** *algorithm, Problem (6) can be globally solved.*

*Proof.* From Algorithm 2, once a support feature $\mathbf{f}_z$ is identified, all corresponding correlated features of $\mathbf{f}_z$ (features with scores $(|c_h| > |c_z| - \sqrt{2\tau}\|\boldsymbol{\alpha}\|)$) will be identified and stored in $\mathcal{Q}$. This fact also implies that all the subsequently selected support features will not be correlated to any of the previously selected support feature. Finally, the top scoring features that satisfy the constraint in (2) then form the support features, hence $\max_{\boldsymbol{\delta}, \mathbf{f}_j \notin \mathbf{X}_{\mathcal{Q}}} \frac{1}{2}\|\sum_{i=1}^{n} \alpha_i y_i (\mathbf{x}_i \odot \boldsymbol{\delta})\|^2$ will be maximized. Inductively, it becomes possible to conclude that the proposed CRM algorithm can solve (6) globally and exactly. This completes the proof. □

The following theorem indicates that the proposed algorithm can globally converge and exhibits the non-monotonic property for feature selection.

**Theorem 2.** *Given that in each iteration of Algorithm 1, the reduced minimax subproblem (5) and the most active constraint selection (6) can be exactly and globally solved, Algorithm 1 stops after a finite number of iterations with a global solution of (4).*

The proof can be adapted from (Tan et al., 2010).

### 3.5. Complexity Analysis

As finding the most violated $\boldsymbol{\delta}$ can be obtained exactly via the Correlation Redundancy Matching algorithm, which firstly sorts the $m$ features, followed by scanning of the $B$ features and then computes the PCC w.r.t. the other features. Here, sorting takes $O(m \log m)$ and finding the support and affiliated features consumes $O(Bmn)$. Therefore, with $T$ iterations on hand, the overall time complexity for CRM is $O(T(m \log m + Bmn))$. As there are at most $TB$ selected features, the training time complexity to solve (5) is $O(TBn)$. Note, by taking benefits from the cutting plane strategy, a relative small $T$ is needed for convergence: as it is noted to converge well within 10 iterations in the experimental studies.

## 4. Experiments

In this section, we conduct experiments to study the feature selection performances of several state-of-the-art methods, including: 1) ReliefF (Robnik-Sikonja & Kononenko, 2003), 2) mRMR[2] (Peng et al., 2005), 3) FCBF[3] (Yu & Liu, 2003), 4) RCFS (Zhou et al., 2010), 5) SVM-RFE (Guyon & Elisseeff, 2003), 6) L1-SVM[4] (Yuan et al., 2011), 7) FGM[5] (Tan et al., 2010), and 8) our proposed GDM using **only support features** for prediction. The first four algorithms belong to filter methods. SVM-RFE is a wrapper method, while the last three are embedded methods. For fair comparisons, all methods except ReliefF, which is integrated in MATLAB R2011b, are implemented in C++ with MATLAB interface. Moreover, the parameters of these methods are configured as suggested by the respective authors. For ReliefF, we report the best results of with $K \in \{1, 2, 3, \ldots, 100\}$ ($K$NN classifier). Further, $C$ is configured to 1 for all methods except L1-SVM, where $C$ varies with different number of selected features. In the experimental study, we set our $\tau = 0.25$ and consider 10, 20, ..., 200 features for each method and report the corresponding resultant training time. Therefore, to facilitate a fair comparison, standard SVM classifier is used to judge the accuracy according to the number of se-

[2]http://penglab.janelia.org/proj/mRMR.
[3]http://www.cs.man.ac.uk/~gbrown/fstoolbox.
[4]http://www.csie.ntu.edu.tw/~cjlin/liblinear.
[5]http://c2inet.sce.ntu.edu.sg/Mingkui/FGM.htm.

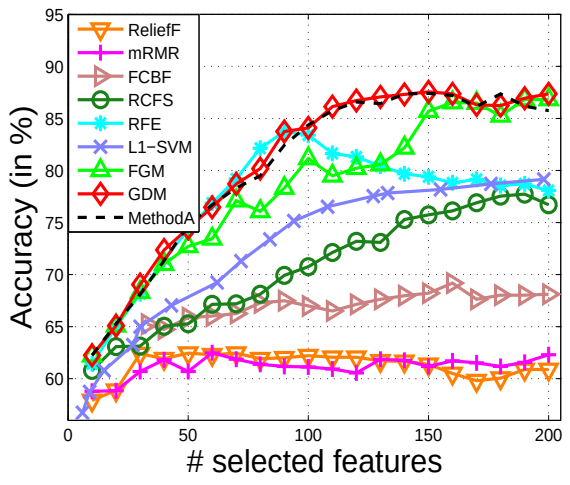

(a) Testing accuracy (in %)

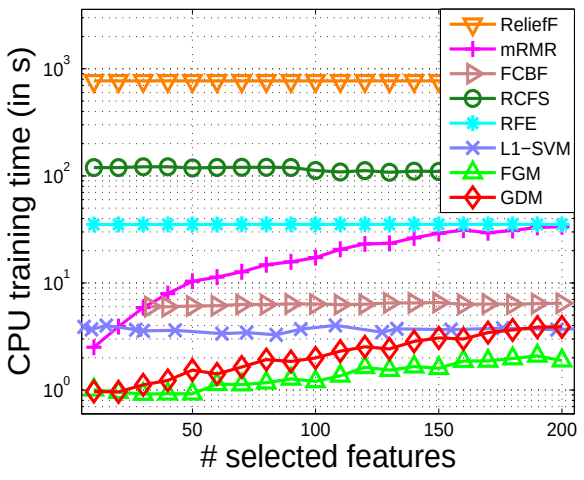

(b) CPU training time (in s)

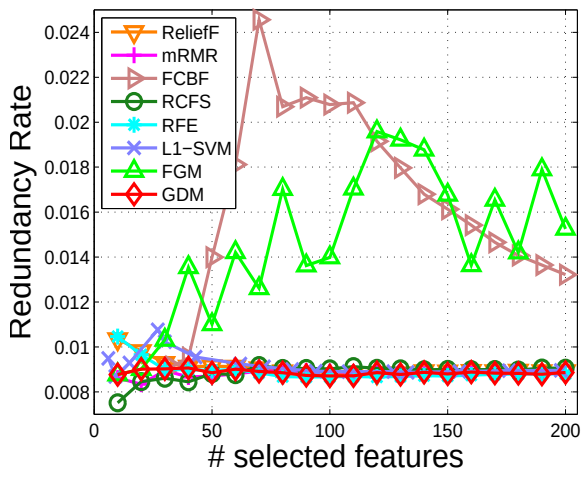

(c) Redundancy Rate

*Figure 2.* Results on Synthetic Dataset.

lected features and their indexes. Further, all experiments are conducted on a PC with Intel® Core™ i7 Processor and 24.0GB memory under Windows Server® 2008.

To evaluate the feature selection performances, three criteria, namely, 1) **Classification Accuracy**, 2) **Training Time Complexity** and 3) **Redundancy Rate** are considered. Following the consistent definition of (Zhao et al., 2012), and assuming $\mathbf{F}$ is the set of selected feature subset with size $\hat{m}$, the redundancy rate can be measured by $\textbf{RED}(\mathbf{F}) = \frac{1}{\hat{m}(\hat{m}-1)} \sum_{\mathbf{f}_i, \mathbf{f}_j \in \mathbf{F}, i>j} |\rho(\mathbf{f}_i, \mathbf{f}_j)|$. The metric assesses the averaged correlation among all selected feature pairs. Obviously, for the same accuracy level, a smaller redundancy rate is deem to be more meaningful.

### 4.1. Evaluation on Synthetic Data

To illustrate the mechanisms of the proposed method, we first conduct a study on a synthetic dataset where the ground truth correlated features are known in advance. The data contains 2,048 observations and 10,000 features, and the predefined predictive features are categorized as 200 feature groups of different sizes, others then serve as noise. Moreover, each of the 30 out of 200 feature groups contains some inner highly correlated features. While in the remaining 170 groups, there is only one feature per group. The predictive ability of each group follows a normal distribution $N(0, 1)$. Noticed that the way of constructing testing set is consistent with training. Our goal is to assess whether a method can correctly identify the relevant members of the ground truth feature set. Nevertheless, for this synthetic data, when the correlated features are truly redundant in the data, removing it would lead the classifier to achieve high accuracy performances. To this end, we manually remove the redundant features directly from the data, (i.e., based on the ground truth available) to solve (6). So that correlation

*Table 1.* Summary of the benchmark datasets.

| Dataset | # Features | Training Size | Testing Size |
|---|---|---|---|
| Mnist | 752 | 11,982 | 1,984 |
| Usps | 676 | 266,079 | 75,383 |
| Kdd2010 | 2,990,384 | 100,000 | 748,401 |
| Webspam | 8,355,099 | 80,000 | 70,000 |

constraint on $\rho(\mathbf{f}_i, \mathbf{f}_j)$ has been imposed. Then we name it as "MethodA".

The experimental results obtained from the synthetic dataset are presented in Figure 2, wherein dash line represented the "MethodA". As expected, we observe from Figure 2(a) that the proposed method outperforms others in most cases; while filter methods such as ReliefF and mRMR achieve the worst prediction performances. Among the filter methods considered, RCFS achieve the highest accuracy since the use of feature clustering helped remove redundancy among features. However, since filter method such as RCFS does not take the classifier into consideration in the feature selection process, the classification accuracy is generally lower than the wrapper and embedded counterparts as expected. FGM outperforms L1-SVM but is noted to be competitive to SVM-RFE when small portion of the features are selected. However, due to the non-convexity optimization formulation of SVM-RFE, which suffers from correlated features, it underperforms FGM when large amount features are considered. Moreover, as FGM imposes a tight convex approximation in the $\ell_0$-model (Tan et al., 2010), it can be observed from Figure 2(a) that both FGM and GDM achieve competitive accuracy result when the number of selected features approaches the ground truth. Thus, it is possible to conclude that the potential of the proposed method can correctly identify the ground truth feature groups.

The training time incurred by all the methods considered is also reported in Figure 2(b). It can be observed that FGM emerges as the fastest among all, while ReliefF incurs the highest training effort. Due to the high cost of feature clustering involving 10,000 features, RCFS also consumes significant training time. Though the proposed GDM incurs slightly more time than FGM, it is noted to be more efficient than the state-of-the-art L1-SVM. In addition, the results on metric **Redundancy Rate** are depicted in Figure 2(c). Overall, GDM achieves competitive low redundancy rate and superior accuracy performance on the synthetic problem considered.

### 4.2. Evaluation on Real-world Data

To assess the practical performance of all feature selection methods, we include a variety of datasets, in both data scale and dimension, which consist of two digit recognition datasets: `mnist`[7], `usps`[6] and two other very high dimensional datasets. The first one is the challenge dataset

[6]http://c2inet.sce.ntu.edu.sg/ivor/cvm.html.

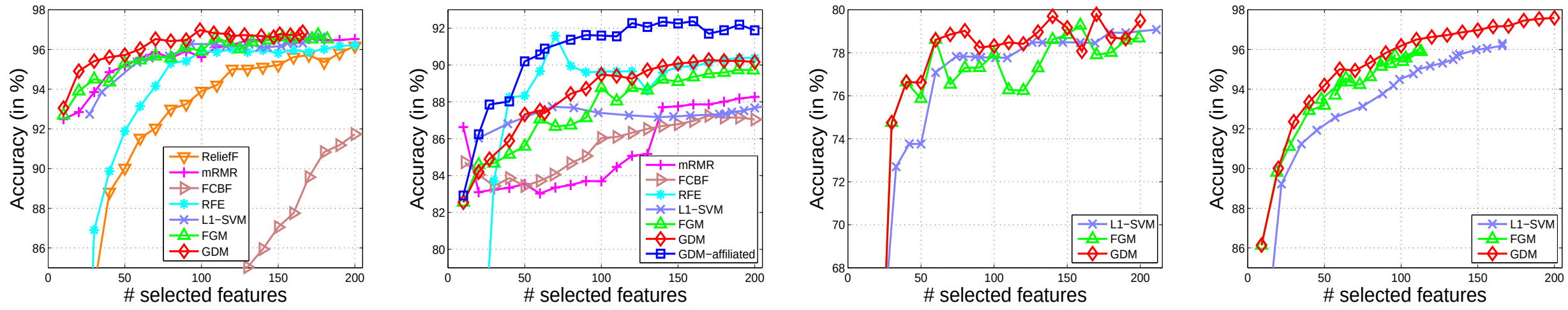

(a) mnist (b) usps (c) kdd2010 (d) webspam

*Figure 3.* Testing accuracy (in %) on Real-world datasets.

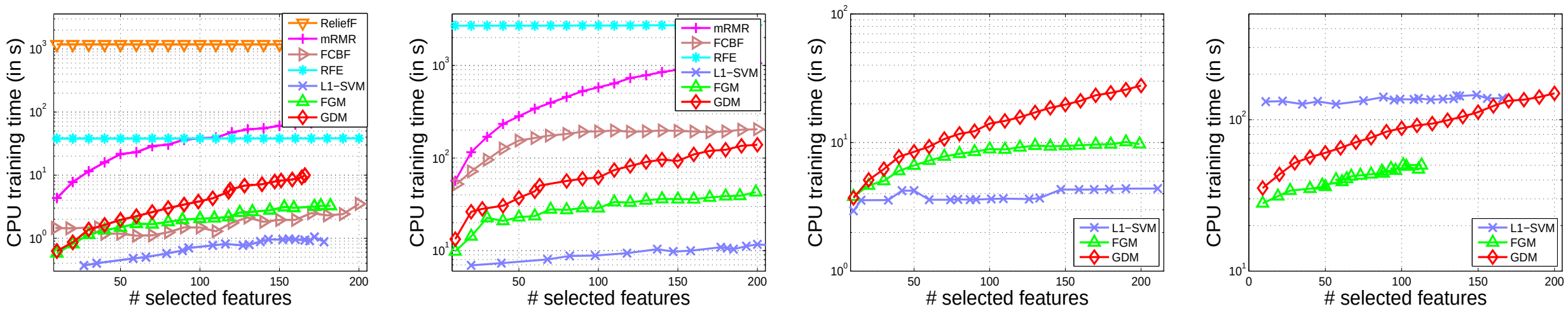

(a) mnist (b) usps (c) kdd2010 (d) webspam

*Figure 4.* Training time (in seconds) on Real-world datasets. (in logarithm scale)

kdd2010[7] used in the educational data mining competition and the second is the spam webpage data webspam[7]. Detailed information about the datasets is listed in Table 1. For kdd2010, we use 10,000 points as the training set and maintain the original testing set. On the webspam data, we randomly select 80,000 points as the training set, while 70,000 for testing. To provide further evidence on the robustness performance of each method, we set 50% as the minimum for accuracy expectation and 1 hour as the maximum training time for all the experiments.

Figure 3 summarizes the accuracy performance attained by the various methods. By appropriately identifying the support feature set, GDM obtains superb accuracy improvements on mnist, kdd2010, and webspam. On the other hand, SVM-RFE performs well only for small number of selected features while deteriorating hereafter as observed in Figure 3(b). When the correlated features are retained to form the affiliated feature groups, the performance of GDM-affiliated is noted to emerge as superior to all the other methods considered. In particular, the results on digit identification in Figure 6 highlights the significance of affiliated feature set, where a comprehensive explanation shall follows later in the discussion and conclusion section. Further, as shown in Figure 5, GDM achieves relatively low redundancy rate in most cases. Although, FCBF and SVM-RFE exhibits lower redundancy rate than GDM on the mnist dataset, the accuracies have been impeded by their high sensitivity to the noise features, as observed from Figure 3(a). Compared to embedded methods, GDM outperforms over FGM and L1-SVM in terms of redundancy reduction, thus exhibiting the improved accuracy. This also implies that GDM can identify a *good feature subset* (Hall, 1999). Since the kdd2010 data is sparse, Figure 4 shows that L1-SVM takes advantage of the sparseness in the data to achieve the shortest training time observed. However, on the webspam data which has more than 8 million features, GDM is noted to learn faster than L1-SVM.

Moreover, there exists the situation that other methods fail in handling the data. For example, RCFS is sensitive to the number of points, thus even on the small mnist dataset with only 11,982 points, the method fails to perform well. SVM-RFE could not maintain an average accuracy of 50% on kdd2010. Both ReliefF and SVM-RFE fail to converge on webspam under the 1 hour maximum training time budget. Moreover, since no filter methods can handle very high dimensional data of webspam and kdd2010, the contests only hold among L1-SVM, FGM and GDM.

## 5. Discussion and Conclusion

In this paper, we have presented a comprehensive study on potential correlated features, leading to the concepts of support feature and affiliated feature. While, superior prediction performance is attained through support features, maintaining some feature redundancies as affiliated features, can be useful for enhanced interpretation of the learning tasks while improving prediction robustness. By taking advantage of the cutting plane strategy, the proposed GDM can handle very high dimensional problems in an efficient way. Notably, the affiliated features are constructed in the proposed method without any additional cost, since they are generated along with the support features.

[7]http://www.csie.ntu.edu.tw/~cjlin/libsvmtools/datasets. As to mnist dataset, class “3” and “8” are taken to form a binary classification problem.

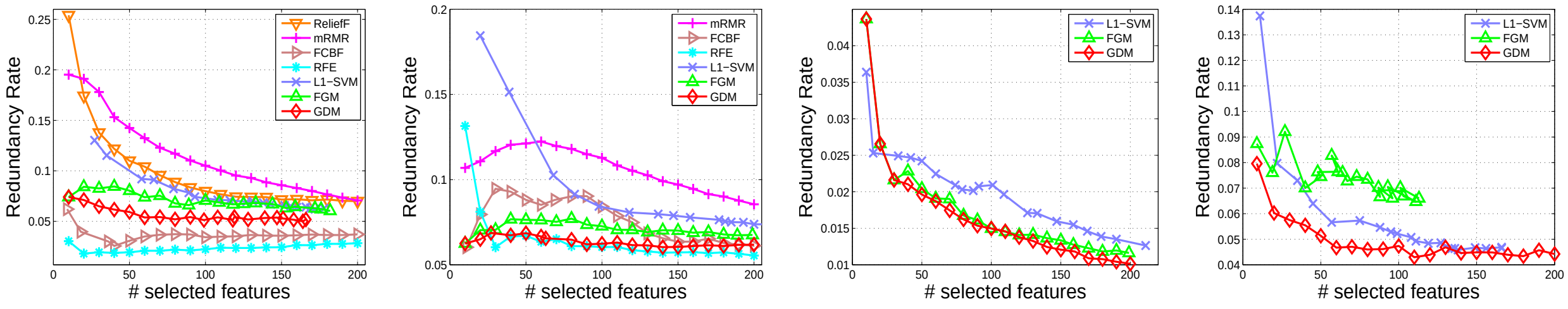

(a) mnist (b) usps (c) kdd2010 (d) webspam

*Figure 5.* Redundancy Rate metric results of various methods.

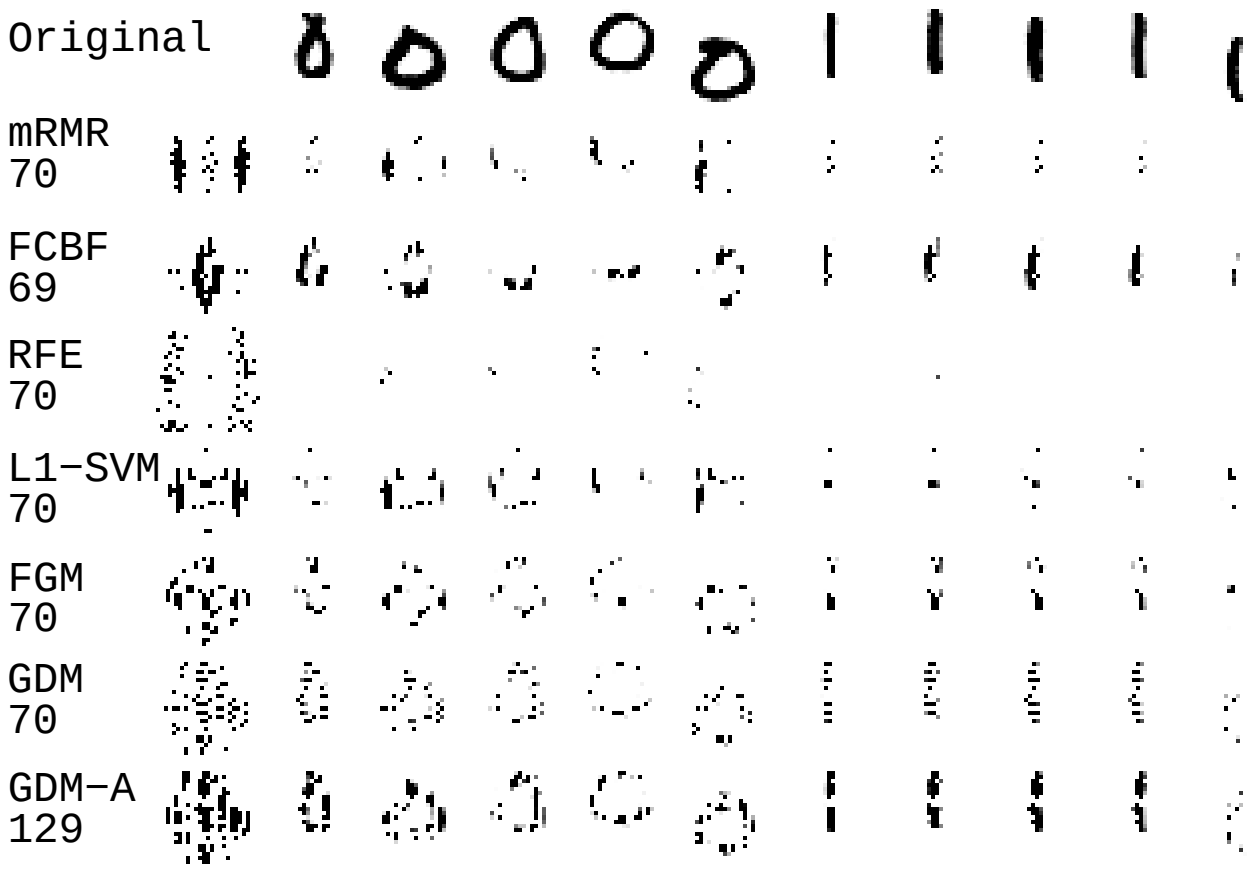

*Figure 6.* Digit identification results of various methods. Example and extracted images by different feature selection methods on `usps` dataset. (Numbers below method indicate # selected features and the adjacent icons are the overall extracted results.)

In what follows, we conclude with further details on the interpretation of the proposed GDM algorithm along with the **affiliated features** attained in Figure 6. With respect to the digit identification result of `usps`[8], the regions highlighted by the affiliated features (129 features) can be useful in assisting the human user in identifying a “0” or “1” from the extracted images of the original pictures. This is consistent to the observation discussed earlier in Figure 1, where the feature groups congregate in the regions of the beard, mustache and silhouette of the face to form the affiliated feature groups. Information with great significance are reserved for further processing. Referring to the affiliated features in Figure 6, despite the highest accuracy achieved by SVM-RFE (70 features) as shown in Figure 3(b), the pixels selected correspond only to the background of the image rather than the digits, other methods also cannot manifest the clear structure of entire digits well. To summarize, we have introduced the notion and significance of correlated features, namely the affiliated feature group in the present paper, and have showcased how it can benefits the task of feature selection in general. We aspire to explore novel constraints that are suitable for the discovery of new structures in high dimensional tasks.

[8]Identified digital handwritten characters extracted from the images of “0” and “1” were gathered in the `usps` data.